\documentclass[runningheads]{llncs}
\usepackage[T1]{fontenc}
\usepackage{xcolor}
\usepackage{amssymb}
\usepackage{amsmath}
\usepackage{graphicx,verbatim}
\usepackage{hyperref}
\usepackage{siunitx}
\usepackage{bbm}
\usepackage{textcomp,gensymb}
\usepackage{float}
\usepackage{makecell}
\usepackage{booktabs}
\usepackage{epstopdf}
\usepackage[utf8]{inputenc}  

\begin{document}
\title{Robust Fetal Pose Estimation across Gestational Ages via Cross-Population Augmentation} 
\titlerunning{Robust Fetal Pose Estimation}
\author{
Sebastian Diaz\inst{1,2} \and
Benjamin Billot\inst{3} \and
Neel Dey\inst{1,4,5} \and
Molin Zhang\inst{2} \and
Esra Abaci Turk\inst{4,6} \and
P. Ellen Grant\inst{4,6} \and
Polina Golland\inst{1} \and
Elfar Adalsteinsson\inst{2}
}
\authorrunning{S. Diaz et al.}
\institute{
Computer Science \& Artificial Intelligence Laboratory, MIT, Cambridge, USA \and
Research Laboratory of Electronics, MIT, Cambridge, USA \and
Inria, Epione team, Sophia-Antipolis, France \and
Harvard Medical School, Boston, USA \and
Massachusetts General Hospital, Boston, USA \and
Boston Children's Hospital, Boston, USA \\
\email{sebodiaz@csail.mit.edu}}

\maketitle
\begin{abstract}
Fetal motion is a critical indicator of neurological development and intrauterine health, yet its quantification remains challenging, particularly at earlier gestational ages (GA). Current methods track fetal motion by predicting the location of annotated landmarks on 3D echo planar imaging (EPI) time-series, primarily in third-trimester fetuses. The predicted landmarks enable  simplification of the fetal body for downstream analysis. While these methods perform well within their training age distribution, they consistently fail to generalize to early GAs due to significant anatomical changes in both mother and fetus across gestation, as well as the difficulty of obtaining annotated early GA EPI data. In this work, we develop a cross-population data augmentation framework that enables pose estimation models to robustly generalize to younger GA clinical cohorts using only annotated images from older GA cohorts. Specifically, we introduce a fetal-specific augmentation strategy that simulates the distinct intrauterine environment and fetal positioning of early GAs. Our experiments find that cross-population augmentation yields reduced variability and significant improvements across both older GA and challenging early GA cases. By enabling more reliable pose estimation across gestation, our work potentially facilitates early clinical detection and intervention in challenging 4D fetal imaging settings. Code is available at \href{https://github.com/sebodiaz/cross-population-pose}{https://github.com/sebodiaz/cross-population-pose}.%
\end{abstract}

\section{Introduction}
Fetal motion is a critical biomarker for neurological development, with diminished movement patterns linked to neurodevelopmental disorders~\cite{deVries08} and intrauterine complications such as hypoxia, infection, and growth restriction~\cite{ayala24}. Accurate quantification of these movements is essential for effective monitoring and timely clinical interventions. Moreover, distinguishing patterns in fetal motion—such as frequency, rhythm, or complexity—may help differentiate between healthy and at-risk populations, offering deeper insight into developmental trajectories.
Traditionally, fetal motion has been assessed through subjective maternal perception, with quantitative automated fetal tracking research still in its infancy. 

Recent methods have used ultrasound for movement quantification by tracking anatomical landmarks~\cite{chen24}, but its associated low tissue contrast makes these landmarks difficult to detect. Fetal Magnetic Resonance Imaging (MRI) is a promising alternative as it enables rapid acquisitions with large field-of-view (FOV) and sufficient contrast to reliably identify anatomical landmarks and enable motion studies.

\begin{figure}[t!]
   \centering
   \includegraphics[width=1.0\textwidth]{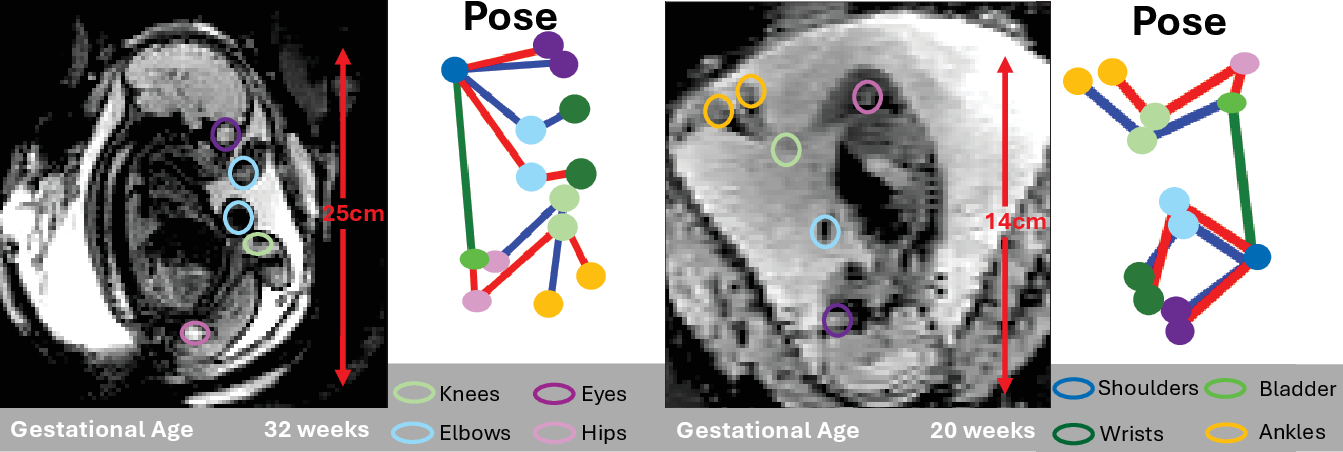}
     \caption{\textbf{A comparison of 3D MRI of fetuses at different gestational ages (GA).} Annotations mark anatomical keypoints. 
     (\textbf{Left}) A 32-week GA fetus scanned at 3mm isotropic resolution, similar to subjects used for training~\cite{xu19}. 
     (\textbf{Right}) A 20-week GA fetus scanned at 2mm isotropic resolution. Pose estimation networks trained on higher GA fetuses (\textbf{Left}) do not generalize well to younger GA fetuses (\textbf{Right}).
     }
   \label{fig:ga_comparison}
\end{figure}

Recent research has demonstrated the potential of convolutional neural networks (CNNs) for fetal pose estimation and motion analysis~\cite{vasung22,xu19}. Inspired by human pose estimation~\cite{Toshev2014,zheng23}, these models predict Gaussian-distributed heatmaps centered on annotated keypoints. The methods achieve impressive performance on third-trimester populations (27-37 weeks GA), but unfortunately this cohort is not fully representative of clinical practice, where scans cover a broader GA range and may have anisotropic resolution (Fig.~\ref{fig:ga_comparison}).
Pose estimation in younger fetuses (18-22 weeks GA) presents further challenges. Compared to third-trimester fetuses, they exhibit less defined joint articulation~\cite{Chauvin2020}, have four times less mass~\cite{Kiserud2017}, and display greater intrauterine mobility, leading to a wider range of poses and frame-to-frame motion. These factors complicate annotation, even for experts, make labeling more time-consuming than their older counterparts, where anatomical features are more distinct and movement is reduced.

Such domain gaps~\cite{pan10} are common in medical imaging due to variations in acquisition protocols, scanner hardware, image resolution, and sites~\cite{Perone2019,Xu2024,Zhang2020}.  Data augmentation, attempts to bridge this gap by perturbing the source domain to encourage robust feature learning~\cite{Yuan2024,Zhang2017}. However, even with the extensive use of existing augmentations, these models fail to generalize to early GAs as they do not fully account for the drastic morphological differences across gestation.

Clinical fetal MRI presents a severe case of domain shift. Fetal growth introduces a fundamental mismatch -- smaller, younger fetuses differ significantly in size, musculoskeletal development, and movement patterns. Without explicitly addressing this discrepancy, models trained on readily available older GA data struggle to generalize to clinical cases at earlier GAs, where accurate pose estimation is most needed.

To address the scarcity of diverse training data for fetal pose estimation, we developed a novel fetal-specific augmentation strategy that captures the unique spatial arrangement of the uterus earlier in development. By leveraging an annotated dataset with segmented regions for the fetus and uterus, we randomly scale and warp fetal bodies and inpaint them onto uteruses from different subjects. This process simulates the wide variation of spatial configurations seen at younger gestational ages (GAs). Importantly, drawing inspiration from methods like SynthMorph~\cite{Hoffmann2022} and SynthSeg~\cite{billot23}, our approach acknowledges that generating realistic data isn't always necessary to achieve significant performance improvements. Through training with our proposed augmentation, we observe substantial gains in fetal pose estimation performance and robustness across all anatomical regions and gestational ages, particularly in clinical cohorts where younger GAs are more common.
\section{Methods}

\subsection{Fetal motion estimation by supervised keypoint detection}

\subsubsection{Problem formulation:}
\begin{figure}[!t]
  \centering
  \includegraphics[width=0.9\textwidth]{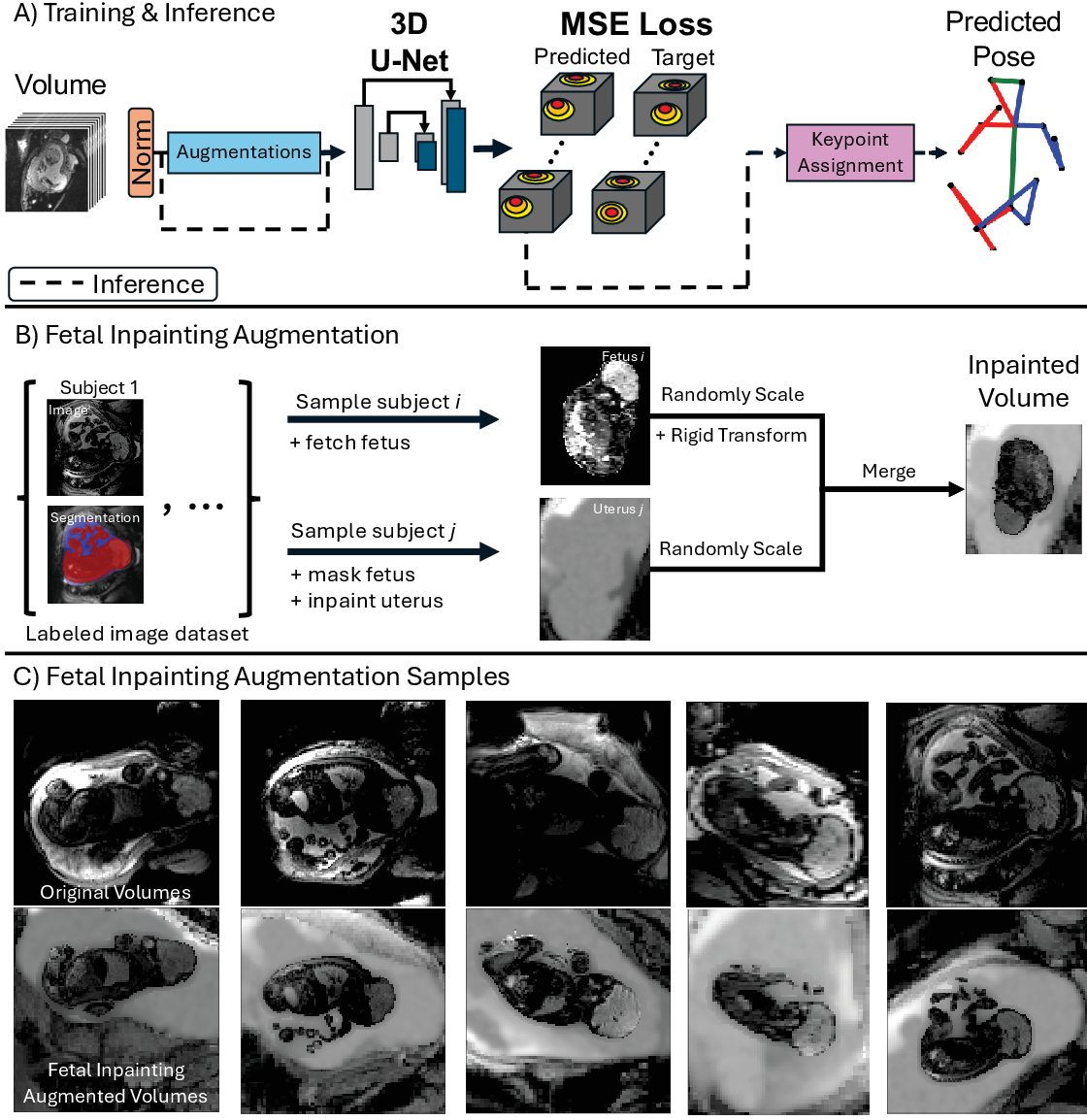}
    \caption{Overview of the proposed method. (A) A network is trained to output Gaussian heatmaps, from which keypoints are extracted by taking the center-of-mass around the heatmap's \texttt{argmax}. (B) Proposed fetal inpainting augmentation. (C) Views of a training volume augmented with the proposed synthetic amniotic fluid and inpainting transform. Top and bottom rows are the raw EPI volumes and associated fetal body embedded into a uterus, respectively. Gamma correction was applied for visualization. 
    }
  \label{fig:setup}
\end{figure}

An overview of our method is seen in Fig~\ref{fig:setup}A. Given an EPI time-series $\mathcal{S} \in \mathbb{R}^{H \times W \times D \times T}$, where $T$ is the number of volumes in the sequence, we define a single 3D volume $V_{t} \in \mathbb{R}^{H \times W \times D}$ as a sample at time $t$ from $\mathcal{S}$. The goal of fetal pose estimation is to localize anatomical landmarks by estimating a set of keypoints
$
{k_{1},k_{2},\dots,k_{15}}
$
where each keypoint $k_{i}$ corresponds to an anatomical landmark in the fetus represented by its 3D voxel coordinates $k_{i} = (x_{i}, y_{i}, z_{i}) \in \mathbb{R}^3$ within the volume $V_{t}$. Instead of directly predicting keypoint coordinates, we represent them with heatmaps $H_{i} \in \mathbb{R}^{H \times W \times D}$, where each voxel $(x,y,z)$ encodes the likelihood of the keypoint location. Specifically, we adopt a probabilistic approach, where the ground truth heatmap for keypoint $k_{i}$ is modeled as a 3D isotropic Gaussian with standard deviation $\sigma$ centered on $(x_{i},y_{i},z_{i})$.

During training, we optimize our network to predict a set of heatmaps $\hat{H} = \{\hat{H}_{1}, \hat{H}_{2}, \dots, \hat{H}_{15}\}$ from an input volume $V$ where $\hat{H}_{i}$ approximates $H_{i}$. The training objective is to minimize the Mean Squared Error (MSE) between the predicted and the ground truth heatmaps. At inference time, the keypoint coordinates, $\hat{k}_{i}=(\hat{x}_{i}, \hat{y}_{i}, \hat{z}_{i})$ are extracted from each predicted heatmap $\hat{H}_{i}$ by first taking the highest activation:
$$
{u}^{*} = (x^{*}_{i},y^{*}_{i},z^{*}_{i}) = \arg \max_{(x,y,z)} \hat{H}_{i}(x,y,z),
$$
followed by a weighted local refinement around the neighborhood, $\mathcal{N} = \{ {u}^{*} + {v}: {v}\in\{-1,0,1\}^{3}\}$, with a small constant, $\epsilon = 10^{-10}$:
$$
\hat{k}_{i} = \frac{\sum_{{u} \in \mathcal{N}}{u}\cdot\hat{H}_{i}({u})}{\epsilon + \sum_{{u}\in \mathcal{N}}\hat{H}_{i}({u})}
$$

\subsection{Fetal-specific augmentations}
\subsubsection{MRI-specific augmentations}
Conventional data augmentation methods -- such as simple rotations, scalings, flips, or intensity adjustments -- benefit generalization by increasing the diversity of training examples~\cite{Isensee2020}. Although these were employed in prior work~\cite{xu19,diffusion_pose,Zhang2020}, they fail to capture the nature of clinical data. 
Building off the augmentations included by the prior methods, we first incorporate a range of transformations to further enhance model robustness. These include additive noise, anisotropy, bias fields, gamma, and spikes typically encountered in MRI acquisitions. In our work, we extend these transformations by increasing their severity over commonly used settings~\cite{billot23,torchio}. 

\subsubsection{Fetal inpainting} To capture the fact that early GA fetuses are smaller, in highly variable poses, and exhibit a substantially higher amniotic fluid-to-body ratio (Fig.~\ref{fig:ga_comparison}), we propose a novel augmentation. Specifically, we extract segmentation masks for the fetal body $B$ and the surrounding amniotic fluid $A$, where the uterus is defined as $U = B \cup A$ and $I$ is original image. We then generate a semi-realistic amniotic fluid intensity distribution in the absence of the fetal body. We achieve this by removing the body and synthesizing the fluid intensities in the vacated space as: 
$I_{\text{s}}(x) = \tilde{A} + \epsilon(x),  x \in B,$
where $\tilde{A}$ is the median of the amniotic fluid intensity and $\epsilon(x)\sim \mathcal{N}(0,\sigma^{2}_{\epsilon})$ represents additive noise. To ensure smooth transitions between the synthesized fluid and the original fluid, we apply a Gaussian kernel $\mathcal{G}(x;\sigma_{u})$ over $U$:
$$
I^{'}(x) = \left[ \mathbbm{1} _{x\notin B}I(x) + \mathbbm{1} _{x\in B}I_{\text{s}}(x)\right] * \mathcal{G}(x;\sigma_{u}), \quad x \in U,
$$
and scale $I^{'}$ by a factor $\gamma \sim p(\gamma)$ resulting in $I_{U}$. We add the newly created image pair $(U, I_{U})$ and body pair ($B$, $I_{B}$) to the training collection (Fig~\ref{fig:setup}B).  To generate the augmentation, a fetal body $B$ with its intensity image $I_{B}$ and uterus image $I_{U}$ with its mask $U$ are randomly sampled from the training collection. We scale $B$ and by a factor $\alpha \sim p(\alpha)$ and apply a random rigid transformation $T \sim p(T)$ to obtain a transformed body $I_{B}(T(\alpha B))$ that satisfies $T(\alpha B) \subset U$. The augmented sample consists of a fetal body $I_{B}(T(\alpha B))$ placed in a randomly sampled synthetic volume $I_{U}$, which is then added to the training dataset.

\subsection{Implementation details}
We perform all augmentations online which increases the variability of the training data and improves robustness of the network. We use TorchIO~\cite{torchio} functions for the conventional augmentation implementations. These augmentations include random noise, random K-space spike artifacts, random bias field corruptions, random rotations, and random gamma adjustments. In the event of a scaling augmentation, we multiply the $\sigma
$ of the ground truth heatmap by the chosen factor to discourage overlapping heatmaps as the subject becomes smaller. We further randomly sample anisotropic image resolution transformations with downsampling factors between 1.5 and 2. Augmentations are not applied to the fetal inpainted volumes. For fetal inpainting subjects, we use an in-house segmentation network, trained on ground truth segmentation masks, to obtain body and amniotic fluid masks.

For keypoint detection, we use a lightweight 3D UNet~\cite{3dunet,unet} variation which has demonstrated superior performance in per-voxel tasks compared to larger models~\cite{anatomix}. Our UNet uses ReLU activations, with 4 pooling operations, an initial embedding dimension of 16, and a channel multiplier of 4, with two convolutional blocks per level. We train our UNet for 1,000 epochs with a batch size of 16, crop size of 64, using Adam optimizer~\cite{adam} with a linear decay scheduler and an initial learning rate of 0.0002.  
\begin{figure}[t!]
  \centering

  \includegraphics[width=\textwidth]{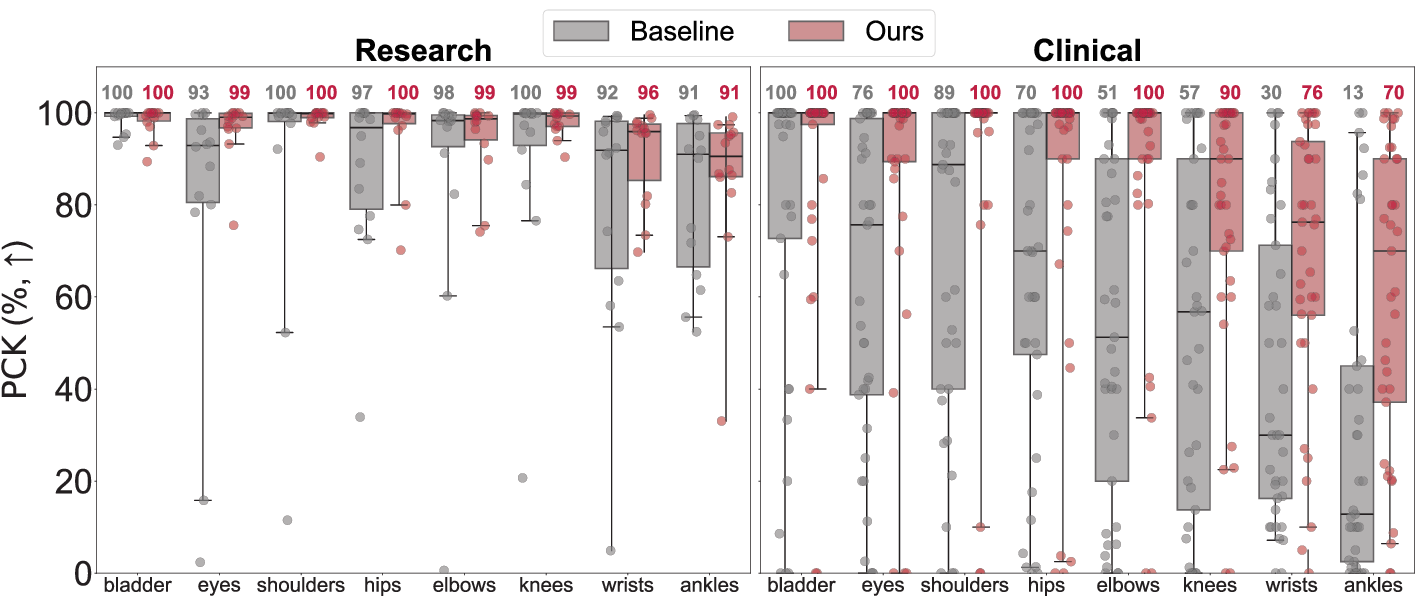}

  \caption{\textbf{PCK (threshold = 10 mm) statistics.} Left and right keypoints are binned together. Median values are reported above each column. Our method demonstrates improvement in the research cohort (left) and significantly improves estimation in the clinical cohort (right).}

  \label{fig:pck_combined}
\end{figure}

\section{Experiments and results}

\subsection{Data}

Our study uses two distinct manually labeled datasets.
The \textbf{\textit{research}} dataset was acquired with a single-shot gradient-echo (GRE) EPI acquisitions, \SI{3}{\milli\meter} isotropic, TR=2.5-4s, TE=32–38ms, FA=90$\degree$. This dataset was collected as part of a study that included predominately higher GAs (27-37 weeks; avg. = 31.74). It includes 19,816 volumes from 77 acquisitions and 53 unique patients. The \textbf{\textit{clinical}} dataset is acquired with the same GRE EPI mentioned above with TR=2.2-5s, TE=37ms, and FA=90\degree. In-plane resolution ranged from {1.8}-\SI{2}{\milli\meter} and slice thickness ranged from {2}-\SI{3}{\milli\meter}. It includes 989 volumes from 37 acquisitions and 28 subjects with an average GA of 21.75 weeks. Thirty of the acquisitions had isotropic \SI{2}{\milli\meter} resolution.
This data was collected during routine clinical acquisitions, making it representative of cases encountered in everyday practice. 
\begin{figure}[!t]
  \centering
  \includegraphics[width=1\textwidth]{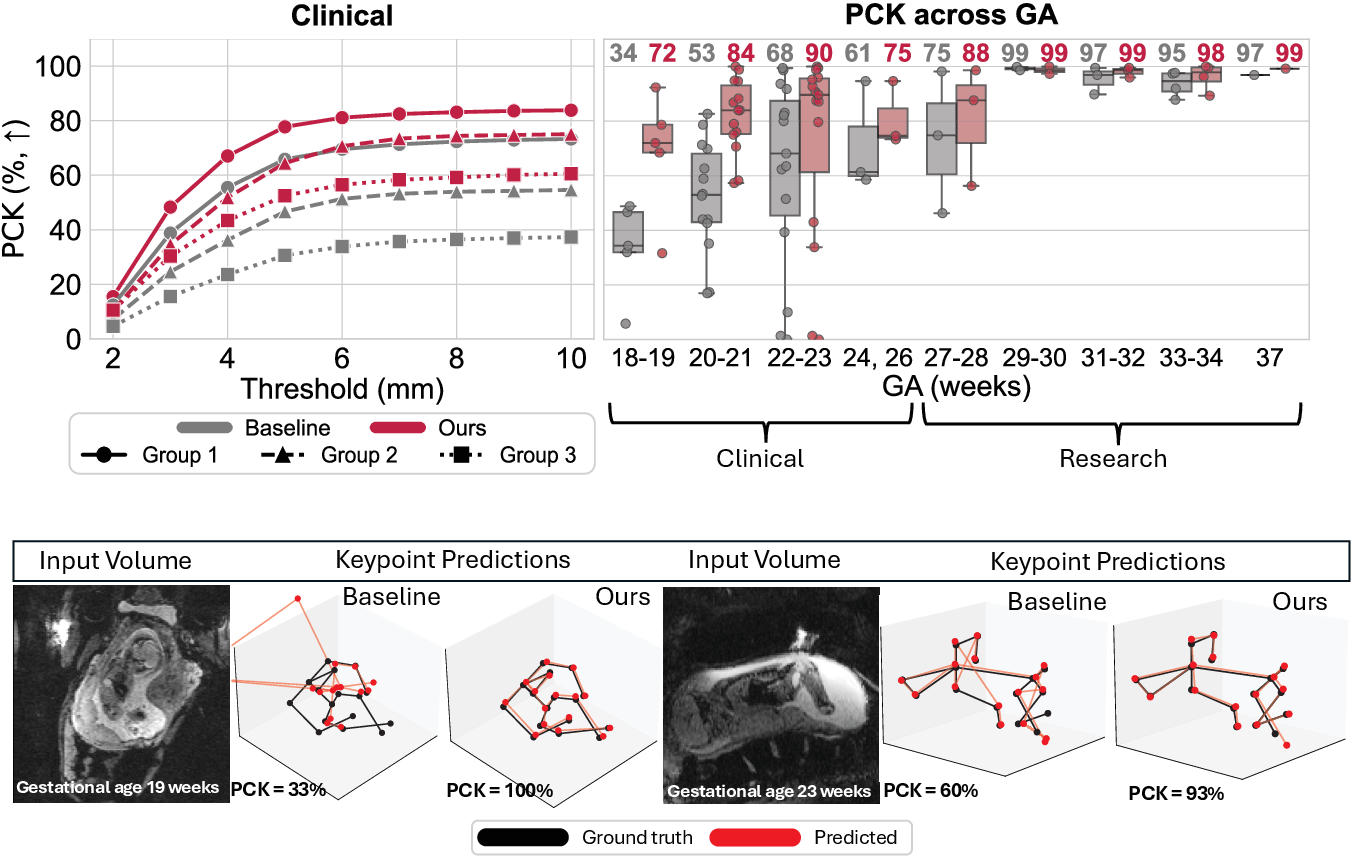}
  \caption{(\textbf{Top Left}) \textbf{PCK vs. Threshold}. Our method versus the baseline's performance on the \textbf{\textit{clinical}}  dataset across threshold values binned into three groups. Group 1 (\textcircled{}): bladder, eyes, shoulders, and hips. Group 2 (${\triangle}$): elbows and knees. Group 3 (${\square}$): ankles and wrists. (\textbf{Top Right}) \textbf{PCK vs. GA}. PCK statistics as a function of GA. Median values are reported above each box. (\textbf{Bottom}) Qualitative predictions. Far right time-series animation is shown in the supplementary.}
  \label{fig:pck_vs_thresh}
\end{figure}

\subsection{Experimental setup}

\subsubsection{Training:} Our networks are exclusively trained on the \textbf{\textit{research}} dataset, dividing (without subject overlap) the 77 times-series into 50 (N=11,618 scans), 13 (N=5,186), and 14 (N=3,462) for training, validation and testing, respectively. 15\% of the total training volumes are fetal inpainting volumes, when applicable. For training, we sampled 8,000 volumes from the training pool.

\subsubsection{Baseline:} We adapt the original method proposed  for volumetric fetal keypoint detection~\cite{xu19} by converting the open-source TensorFlow 1.x~\cite{tensorflow2015-whitepaper} code (\nolinkurl{https://github.com/daviddmc/fetal-pose})
into PyTorch~\cite{pytorch}. This method has demonstrated high performance in research-grade acquisitions~\cite{xu19} and influenced observation motion analysis research~\cite{vasung22}, making it the most relevant and meaningful baseline for comparison. We maintain the same pre-processing, augmentation, architecture, and training parameters, training for 400 epochs using a batch size of 8, a crop size of 64, the AdamW optimizer~\cite{adamw} (weight decay = 0.0001), a cosine warmup scheduler, an initial learning rate of 0.001. Data augmentation settings were replicated from~\cite{xu19} and included intensity, rotations, and scaling transformations.

\subsection{Results}
We evaluate keypoint localization using percentage of correct keypoints (PCK), which is defined as a frequency the predicted keypoint falls within a predefined threshold distance from the ground truth. Figure~\ref{fig:pck_combined} reports PCK statistics across different acquisitions in the research (n=14) and clinical (n=37) datasets. Our method consistently outperforms the baseline, achieving higher median PCK scores with notably lower variability across acquisitions, highlighting its robustness. In particular, keypoints corresponding to the wrists and ankles -- areas indicative of significant gestational motion -- exhibit median performance increases of 46\% and 57\%, respectively, in the clinical cohort.

We further stratified the keypoints into three groups based on detection difficulty: Group 1 (eyes, shoulders, hips, bladder), Group 2 (elbows, knees), and Group 3 (ankles, wrists). As illustrated in Figure~\ref{fig:pck_vs_thresh}, our method delivers substantial and consistent gains across all groups. The most dramatic improvements are observed in Group 3, which comprises the smallest and most challenging features.

To investigate the influence of GA on detection accuracy, we merged the datasets and performed a PCK analysis within discrete GA bins (Fig.~\ref{fig:pck_vs_thresh}). As expected, performance saturates for both methods on older subjects, while our method demonstrates a significant advantage on younger populations. Furthermore, the narrower spread of errors across subjects underscores the robustness and consistency of our approach.

\begin{table}[!t]
\centering
\setlength{\tabcolsep}{1.25pt}
\scriptsize
\begin{tabular}{lcccccccc}
\toprule
\textbf{Method} & \textbf{Bladder} & \textbf{Eyes} & \textbf{Shoulders} & \textbf{Hips} & \textbf{Elbows} & \textbf{Knees} & \textbf{Wrists} & \textbf{Ankles} \\
\midrule
\multicolumn{9}{c}{\textbf{Research data}} \\
\midrule
Ours & 98 ± 3 & \textbf{97 ± 6} & \textbf{99 ± 2} & \textbf{96 ± 9} & \textbf{95 ± 9} & \textbf{98 ± 3} & \textbf{91 ± 10} & \textbf{86 ± 16} \\
Ours$-$FI & 98 ± 2 & 91 ± 23 & 95 ± 13 & 93 ± 13 & 89 ± 23 & 92 ± 19 & 83 ± 19 & 82 ± 21 \\
Ours$-$FI$-$Aniso. & \textbf{99 ± 2} & 88 ± 22 & 93 ± 19 & 90 ± 19 & 90 ± 21 & 93 ± 19 & 81 ± 25 & 85 ± 14 \\
Ours$-$FI$-$Int. & \textbf{99 ± 2} & 80 ± 32 & 91 ± 24 & 87 ± 26 & 85 ± 29 & 92 ± 18 & 80 ± 26 & 78 ± 21 \\
Ours$-$FI$-$Noise & 97 ± 6 & 87 ± 27 & 92 ± 24 & 94 ± 10 & 86 ± 27 & 90 ± 21 & 83 ± 26 & 79 ± 24 \\
Ours$-$FI$-$Spike & 99 ± 3 & 83 ± 30 & 92 ± 25 & 93 ± 12 & 87 ± 26 & 91 ± 21 & 85 ± 25 & 81 ± 19 \\
Ours$-$FI$-$Scale & 98 ± 3 & 87 ± 25 & 89 ± 27 & 86 ± 26 & 84 ± 27 & 88 ± 26 & 77 ± 27 & 79 ± 25 \\

\midrule
\multicolumn{9}{c}{\textbf{Clinical data}} \\
\midrule
Ours & \textbf{88 ± 25} & \textbf{86 ± 29} & \textbf{90 ± 27} & \textbf{84 ± 32} & \textbf{87 ± 27} & \textbf{77 ± 29} & \textbf{69 ± 29} & \textbf{61 ± 33} \\
Ours$-$FI & 84 ± 28 & 84 ± 28 & 87 ± 26 & 81 ± 31 & 82 ± 27 & 74 ± 35 & 66 ± 32 & 57 ± 34 \\

Ours$-$FI$-$Aniso. & 77 ± 35 & 78 ± 32 & 80 ± 31 & 64 ± 38 & 68 ± 33 & 50 ± 36 & 53 ± 29 & 43 ± 32 \\
Ours$-$FI$-$Int. & 71 ± 38 & 54 ± 39 & 71 ± 35 & 54 ± 38 & 71 ± 30 & 55 ± 37 & 50 ± 34 & 35 ± 30 \\
Ours$-$FI$-$Noise & 76 ± 36 & 85 ± 31 & 86 ± 30 & 69 ± 38 & 81 ± 30 & 70 ± 36 & 68 ± 28 & 49 ± 37 \\
Ours$-$FI$-$Spike & 84 ± 30 & 76 ± 34 & 83 ± 31 & 73 ± 35 & 78 ± 32 & 62 ± 35 & 59 ± 35 & 39 ± 32 \\
Ours$-$FI$-$Scale  & 72 ± 38 & 77 ± 32 & 83 ± 29 & 66 ± 35 & 72 ± 33 & 59 ± 38 & 62 ± 32 & 42 ± 35 \\

\bottomrule
\end{tabular}
\caption{\textbf{Ablation}: PCK (Mean ± standard deviation, threshold = \SI{10}{\milli\meter}) per acquisition on both datasets. \textbf{Bolded} numbers are highest performance in each dataset. "FI" denotes Fetal Inpainting. "Int." is short for "intensity" and refers to gamma and bias field transforms. "Aniso." denotes anisotropy.}
\label{tab:ablation}
\end{table}

\subsubsection{Ablations:}
To understand the effect of inpainting augmentation in conjunction with other standard transformations, we conduct ablations on each augmentation while maintaining the same training hyperparameters and report results in Table~\ref{tab:ablation}. We find that all added augmentation contribute to the robustness of the model across all anatomical landmarks, with fetal inpainting combined with all remaining augmentations yielding the best model overall.

\section{Conclusion}
This study addresses the clinical need for a more reliable fetal pose estimation method. We demonstrate how our augmentation approach, including a fetal inpainting method, can significantly improve performance, particularly in the early GA range. These advancements make the model clinically applicable, enabling clinicians to answer critical questions regarding fetal motion and neurological development with greater accuracy.

    

\begin{credits}
\subsubsection{\ackname} NIH R01 HD114338, R01EB032708, EB036945, 1R01EB036945. MIT SoE Fellowship. MathWorks MATLAB Fellowship.

\subsubsection{\discintname}
The authors have no competing interests to declare that are
relevant to the content of this article.
\end{credits}

%
%
%
\bibliographystyle{splncs04}
\bibliography{bibliography}

\end{document}